# High-Level Description of Robot Architecture

Sabah Al-Fedaghi  
Computer Engineering Department  
Kuwait University  
Kuwait  
Sabah.alfedaghi@ku.edu.kw

Manar AlSaraf  
Computer Engineering Department  
Kuwait University  
Kuwait  
manar.alsaraf@grad.ku.edu.kw

*Abstract*—Architectural description (AD) is the backbone that facilitates the implementation and validation of robotic systems. In general, current high-level ADs reflect great variation and lead to various difficulties, including mixing ADs with implementation issues. They lack the qualities of being systematic and coherent, as well as lacking technical-related forms (e.g., icons of faces, computer screens). Additionally, a variety of languages exist for eliciting requirements, such as object-oriented analysis methods susceptible to inconsistency (e.g., those using multiple diagrams in UML and SysML). In this paper, we orient our research toward a more generic conceptualization of ADs in robotics. We apply a new modeling methodology, namely the thinging machine (TM), to describe the architecture in robotic systems. The focus of such an application is on high-level specification, which is one important aspect for realizing the design and implementation in such systems. TM modeling can be utilized in documentation and communication and as the first step in the system's design phase. Accordingly, sample robot architectures are re-expressed in terms of TM, thus developing (1) a static model that captures the robot's atemporal aspects, (2) a dynamic model that identifies states, and (3) a behavioral model that specifies the chronology of events in the system. This result shows a viable approach in robot modeling that determines a robot system's behavior through its static description.

## I. INTRODUCTION

Robotic systems are multifaceted and challenging. Thus, the robotic systems must interact with a dynamic environment to be reactive and flexible to unexpected changes. Such challenges require good frameworks and models that embody well-defined concepts to effectively manage this complexity. The use of a well-conceived architectural description (AD) can often help to manage that complexity [1]. An AD is a representation of a system, its structure, and associated behaviors, such as the AD languages UML and SysML [2].

An architectural model is the backbone that facilitates the description, implementation, and validation of robotic systems [3]. It is important for communication among stakeholders to provide a common language in which different concerns can be expressed, negotiated, and resolved at a level that is manageable even for complex systems [4]. Additionally, the architecture helps with recognizing constraints, dictating organizational structures, enabling a system's quality attributes, managing changes, and providing the basis for training [4].

This paper applies a new modeling methodology, the thinging machine (TM), for architecting robotic systems. The focus of such an application is on *a high-level AD*, which is one important aspect of designing and implementing a robotic system [4]. The AD can be utilized in documentation and communication and as the first step in the system's design.

## II. RELATED WORKS

Robot architecture is a subtopic of system architecture. Robot architecture is hardly recognized as an independent subject. For example, a search on "robot architecture" on Wikipedia produces the response, "The page 'Robot architecture' does not exist"; instead, several pages are given, such as "autonomous robot architecture" and "subsumption architecture."

We outline here some of the many sources in the rich field of system architecture, starting with the types of structures in this field. England [2] lists 28 sample architectural domains, including conceptual architecture, computer (hardware) architecture, software architecture, communication architecture, technical architecture, and reference architecture. Architecture-related standards have been adopted to address lifecycle processes, activities, and tasks, such as the IEEE Standard Ontology for Robotics and Automation, IEEE/RS, INCOSE UK's Practice of System Architecture (2014), and ISO/PAS 19450:2015 Automation Systems and Integration—Object-Process Methodology. A survey reported a list of 120+ AD languages, which are detailed in [2]. Because of space limitations, we focus on four representative samples of robot ADs.

An architecture comprises the high-level schema that show a system's overall structure [4]. The term refers to "determining the needs of the user of a structure and then designing to meet those needs as effectively as possible within economic and technological constraints… The emphasis in architecture is upon the needs of the user, whereas in engineering the emphasis is upon the needs of the fabricator" [5]. The term architecture is used here to describe the *conceptual structure* and functional behavior, as distinct from the organization of the logical design and the physical implementation [6].

The Software Engineering Institute defines software architecture as a system's structure, which includes system elements, their externally visible interfaces, and the relationships among them in the system [4]. Software architecture deals with an abstraction of a system, by defining how elements interact within this abstraction but not how individual elements are implemented [7]. According to Bass, Clements, and Kazman [4], there is "little difference" between software architecture and system architecture. The architectural view is abstract, distilling implementation details and concentrating on the system elements' behavior and interactions [4]. Architecture prescribes a system's structure by accommodating combinations of both physical structure and functionality (utility) [2].

Coste-Maniere and Simmons [3] assert that the "architectural structure refers to how a system is divided into subsystems, and how those subsystems interact. This is often represented by the traditional 'boxes and arrows' diagrams." If there is no system architecture, the project should not proceed to full-scale system development [8]. According to Coste-Maniere and Simmons [3], a robot system often uses several architecture styles together, so it is sometimes difficult to determine exactly what architecture is used—to describe the robot's system—because the architecture and the implementation are often intimately tied together. Coste-Maniere and Simmons [3] continue, "This is unfortunate, as a well-conceived architecture can have many advantages in the specification, execution, and validation of robot systems."

To exemplify the types of robot AD, we show four representative cases. The purpose is not to give fair accounts of them, but to show the types of diagrams used for those cases for contrast with the TM diagrams developed later in the paper.

Loza-Matovelle, Verdugo, Zalama, and Gómez-García-Bermejo [9] developed a system that combines robots with a network of sensors and actuators, as illustrated in Figure 1. Different devices are represented by heterogeneous icons such as a device, a face, a hand, and a telephone. Servers in the system are represented as circles or rounded rectangles. Snoswell et al. [10] presented a robot system architecture (Figure 2) for a manipulator that grasps and completes tasks. The system is distributed across multiple computers. They tested this system architecture using the so-called MOVO mobile manipulation platform from Kinova Robotics.

As shown in Figure 3, PatentSwarm [11] (in an invention) incorporates state machines into the robot AD. Note that these high-level ADs are static depictions that do not incorporate dynamic features or facilitate movement to the next level of development (i.e., design). This is an important point to consider when contrasting them with the proposed TM modeling.

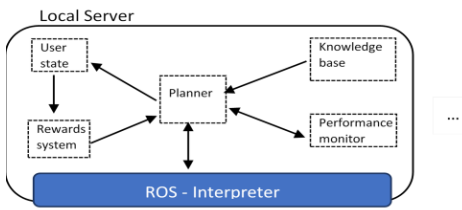

Fig. 1. Sample AD that combines robots with a network of sensors and actuators (adapted from [9]).

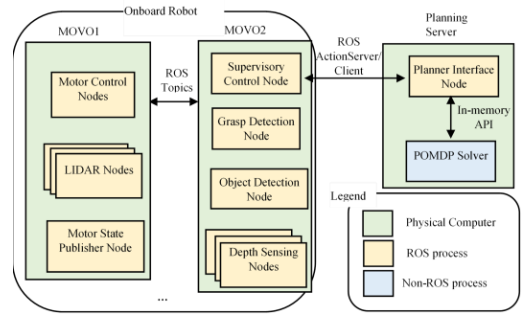

Fig. 2. Architecture for a distributed robot-planning system (adapted from [10]).

The fourth type of AD lacks a holistic view of the robot system. According to Bass, Clements, and Kazman [4], architectural structures can be divided into three groups: module structures, component-and-connector structures, and allocation structures. Module-based structures include decomposition and use layers and classes or generalization. Component-and-connector structures include processes, concurrency, and the client–server structure. Finally, allocation structures include allocation and deployment. Bass, Clements, and Kazman [4] used many diagrams, such as UML diagrams, to describe different aspects of architecture. Figure 4 shows a sample of these diagrams, called data flow architectural views.

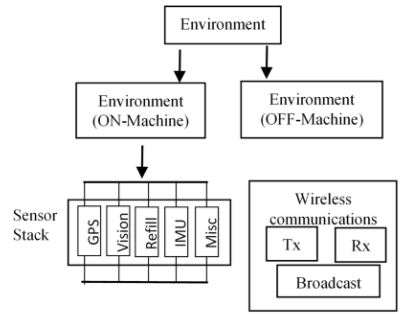

Fig. 3. A schematic illustration of an embodiment of the Robot Control Architecture (partially from [11]).

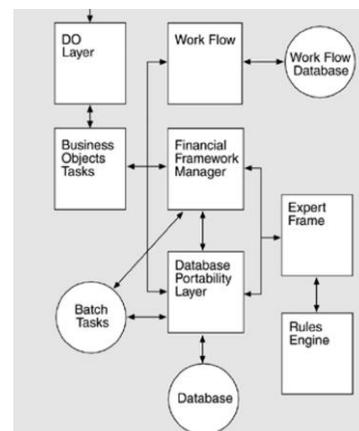

Fig. 4. Data flow architectural view (partially from [4]).

## III. RESEARCH PROBLEM AND PROPOSED SOLUTION

The samples presented in the previous section demonstrate the need for an AD that reduces ambiguity and misunderstandings (e.g., via consistent model usage), manages complexity (e.g., via abstraction, with only the salient features presented), and affords assurance (i.e., correct interpretation) [2]. Bass, Clements, and Kazman's [4] representation of AD systems can be criticized as over-described when using a version of Occam's razor indicating that things should not be multiplied without necessity. In general, current architecture specifications might vary and lead to various difficulties, including mixing architectural specifications with implementation issues. Such descriptions lack the qualities of being systemic and coherent, as well as technical-related forms (vs. drawings of physical layouts and structural compositions in housing). Many system architectures use icons (e.g., faces, hands, computer screens) without a reasonable level of detail. Additionally, a variety of languages exist for eliciting requirements (e.g., object-oriented analyses use scenarios or "use cases" to embody requirements) and finite-state-machine models [4] that are susceptible to inconsistency (e.g., multiplicity of diagrams in UML and SysML). From the modeling point of view, such representations mix static modeling with dynamism that incorporates time. The difference between staticity and dynamism will become clearer when we discuss our method of modeling robot architecture.

On the other hand, for a robot architecture to be effective as the backbone of a project's design, the architecture's documentation should be informative, unambiguous, and readable by many people with various backgrounds [4]. We will show that our TM AD (called the static model, denoted by S) can be specified by a single ontological element called the thimac (*thi*ngs/*ma*chines). S is decomposed to produce sub-diagrams that can be converted to *events* by infusing a time element into the model. The events' chronology models the system's behavior.

Before applying TM to robot architecture specification, the next section provides a summary review of TMs. TM modeling is a promising modeling approach that has been applied in diverse areas such as designing unmanned aerial vehicles [12], documenting computer networks [13], modeling network architectures [14], modeling advanced persistent threats [15], modeling an IP phone communication system [16], and programming [17]. The TM model can also be used to model service-oriented systems [18], business systems [19], a tendering system [20], a robot's architectural structure [21], the VLSI engineering process [22], physical security [23], the privacy of the processing cycle for bank checks [24], a small company process [25], wastewater treatment controls [26], asset-management systems [27], IT processes using Microsoft Orchestrator [28], digital circuits [29], and automobile tracking systems [30].

## IV. THINGING MACHINE MODELING

According to the IEEE-RAS (Robotics and Autonomous Systems) working group on ontologies for robotics and automation, with the growing complexity of behaviors that robots are expected to perform, the need for well-defined knowledge representation is becoming more evident [31]. In this context, ontologies are defined as "[consisting of] a formal conceptualization of the knowledge representation and [providing] the definitions of the concepts and relations capturing the knowledge of a domain in an interoperable way" [32]. Examples of such ontologies include those of Cheng et al. [33]: device (e.g., concepts such as that of a machine), process (e.g., operations performed by technical equipment), parametric (e.g., quality of service), and product ontologies (e.g., product information). Engel, Greiner, and Seifert [34] proposed ontologies for batch process plants that include operations, architectures, and general system characteristics and relations.

In this paper, we orient our research toward a more generic conceptualization of ontologies' role in robotics. An ontology is a crucial mechanism with which to model a robot system and its activities. A model refers to a conceptual description of a robot system and its processes. Developing such a model restrains and guides the robot system's design, development, and use. The issue, in this context, is a cross-area study between modeling and ontology in robotics. This paper provides a broad ontological foundation for conceptual modeling in the robotics domain by suggesting a practical ontology in terms of the notion of TMs. TM modeling uses a one-category ontology called a thimac in contrast to objects, attributes, and relations in the object-oriented paradigm. In philosophy, *tropes* are a well-known one-category ontology. According to Cheng et al. [32], "One-category ontologies are deeply appealing, because their ontological simplicity gives them an unmatched elegance and sparseness."

Let a thimac be denoted by Δ; then, Δ = (M. T), where Δ has a dual mode of being: a machine denoted as M and a thing denoted by T. Figure 5 shows a general form of TM modeling machines, and Figure 6 is a simplification of Figure 5. M includes generic actions described as follows. A sample of the two sides of a thimac will be given later.

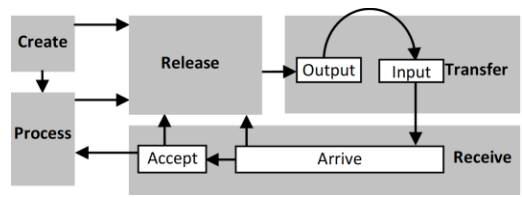

Fig. 5. The thinging machine, M.

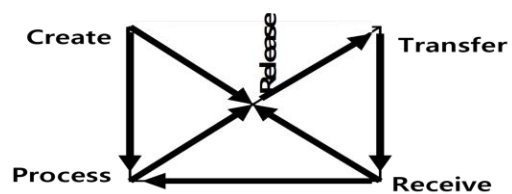

Fig. 6. Simplification of machine, M.

The actions (also called stages) in M (Figure 5) can be described as follows:
- *Arrival*: A thing reaches a new machine.
- *Acceptance*: A thing is allowed to enter the machine. If arriving things are always accepted, then arrival and acceptance can be combined into a "receiving" stage. For simplicity, we will assume that a receiving stage exists.
- *Processing* (alteration): A thing undergoes modifications without creating a new thing.
- *Release*: A thing is marked as ready to be transferred outside of the machine.
- *Transference*: A thing is input or output outside of or within the machine.
- *Creation*: A new thing is born (created) within a machine. Creation can designate bringing into existence (e.g., ∃ in logic) in the system because what exists is what is found. Creation in M indicates "there is" in the system, but not at any particular time.

The TM model also includes the notion of *triggering*, which connects two sub-diagrams between which there is no flow. Triggering is represented by dashed arrows in the TM diagram.

To informally justify the five TM actions, consider a robot's actions. The robot interacts with the environment either through inputting or outputting. Through its interface (*transfer*), it *receives* things (e.g., data or actions) and outputs (transfers) things (e.g., data or sound). Some of these output things might be "stocked" (released), waiting until the right time for output. Accordingly, the transfer, receive, and release actions are all types of interactions with the outside, which are usually referred to as sending data, receiving actions (e.g., physical hits), outputting movement (e.g., walking to a certain position), etc. Additionally, the robot might process incoming things such as converting a signal to data, analyzing a scene, inspecting a sound, and so on. It also could create (generate, produce) things such as a sound, movement, or plan. All activities can be specified in terms of the five actions—create, process, release, transfer, and receive—or a subset of these actions.

## V. EXAMPLE: A WINDOW-OPENING ROBOT

Cassinis [35] developed a robot that, when given the goal "open the window," could perform the following sequence of steps: (1) locate the window, (2) reach near the window, (3) locate the handle, (4) reach the handle, (5) turn the handle, and (6) pull the handle.

### A. The Static TM Model

Figure 7 shows the TM model S of this window-opening task. We assume that the window's location is communicated by a sensor and that the handle position is recognized through a camera on the robot.

Upon being activated to open a window, the robot receives data about the window's location (circle 1 in the figure) and processes the data (2) to trigger (3) the window position's generation (4). The window position and robot's current position (5) both flow (6 and 7) to be processed (8), triggering (9) the creation of a description of the path to reach the window (10). This path flows (11) to the wheel control (12), where the path data are processed (13) to generate movement (14) toward the window. Upon reaching the window (15), two triggering actions occur:
- The robot's new location replaces its current location (16 and 17).
- The camera is turned on to search for the handle (18 and 19).

Upon collecting the data about the handle (20), the handle position is recognized and processed (21). Such a process triggers the creation (22) of the required trajectory to reach the handle (23), which flows (24) to the handle (25). There, the trajectory is processed (26) to trigger the handle's movement (27) to perform the following:
- Turning the handle (28) and
- Pulling it (29).

Note that TMs are applied uniformly for all types of things: data, processes, wheels, handles, camera, movement, pulling, and turning. Every machine is constructed from the create, process, release, transfer, and receive actions, or from a subset of these actions. Model S is richer than the so-called ADs; however, if we are interested in the system in terms of its components and their relations, then these can be extracted from S by eliminating the actions, as shown in Figure 8. Every component in Figure 8 is a thimac. To illustrate the notion of thimac, Figure 9 shows the window-opening robot as a thimac. It is a machine and a thing simultaneously. For example, as a thing, it can include its physical attributes, manufacturer, etc., as in the case of a class's attributes in the object-oriented model. In addition, as a thing, it can be shipped, cleaned, etc.

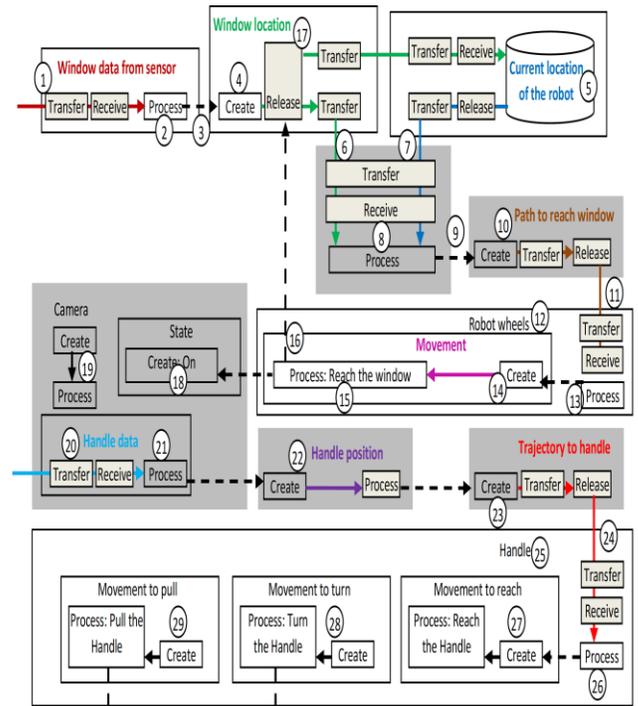

Fig. 7. Static TM Model, S

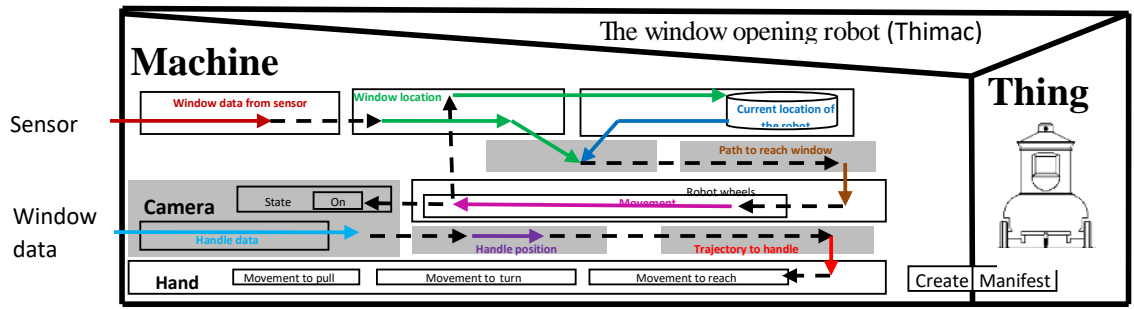

Fig. 8. The window-opening robot as components and their relationships.

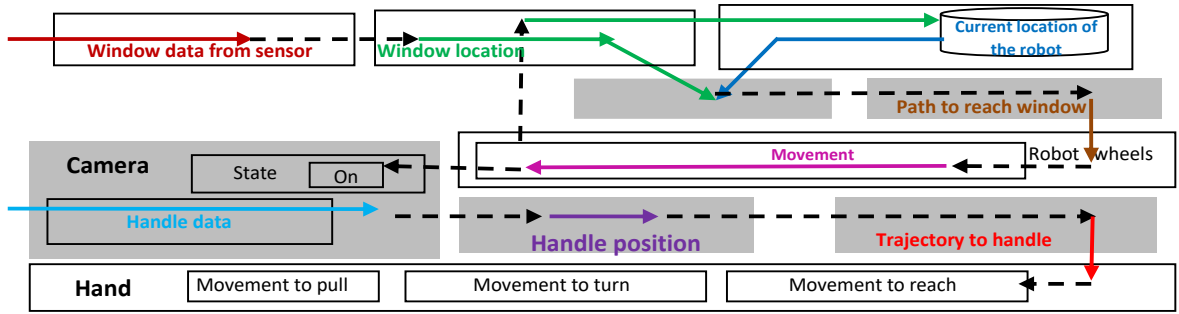

Fig. 9. The window-opening robot as a thimac.

Alternatively, if we want to go in the opposite direction into the model's fine details, we can apply the same TM machine to the subthimacs. Suppose that we add an obstacle in the path to the window. Figure 10 shows the needed modifications to the original model S (Figure 7).

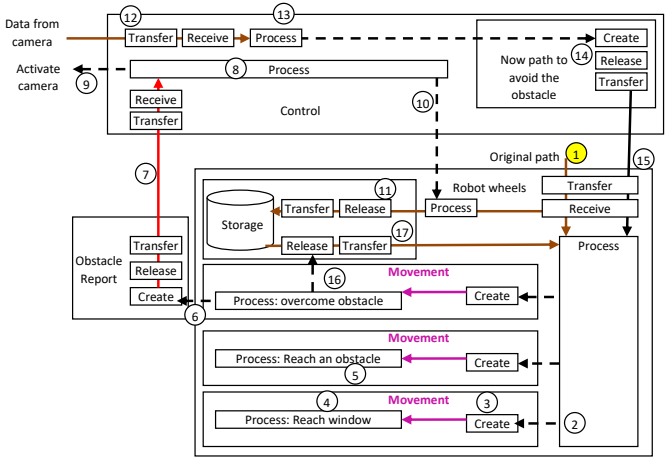

Fig. 10. Adding an obstacle to the window-opening robot.

The modification starts in the wheels machine, where the original path to the window (circle 1) reaches the wheels' controller to be processed (2), triggering a movement (3) to reach the window (4). Suppose the movement instead meets an obstacle (5), which triggers (6) a warning. The warning flows (7) to a control module, which processes it (8) to trigger,

(a) activating a camera on the robot (9) and (b) saving the current path in storage (10 and 11) to continue later after overcoming the obstacle.

The camera data (12) are analyzed (13) to trigger the creation of a new path (14). The new path flows to the wheels system (15), where it is processed to create movement. After the obstacle is overcome (16), the path to the window is restored (17).

*B. The Dynamic Model*

S is a machine schema that can be decomposed to generate a new organizational level (multiplicity) from the "meaningful" parts of S. Model S (Figure 7) is a static description that represents a still or resting (no time) condition. The meaningfulness of a part of S resides in the isomorphism between the part and the thing it is supposed to represent (in the modeler's conceptual framework). Decomposition is necessary because the system described by S is clearly "activated" behaviorally, piece by piece (sub-diagrams). Figure 11 shows a selected division of S for the robot system in 15 static changes. The robot's dynamism originates from *conceptually* dividing it as a whole and replacing it with its 15 sub-diagrams, which then become viewed as events by injecting a time sub-machine into each of them. For example, the event *Replace the old robot location with the current location* is modeled as shown in Figure 12. The chronology of events is shown in Figure 13.

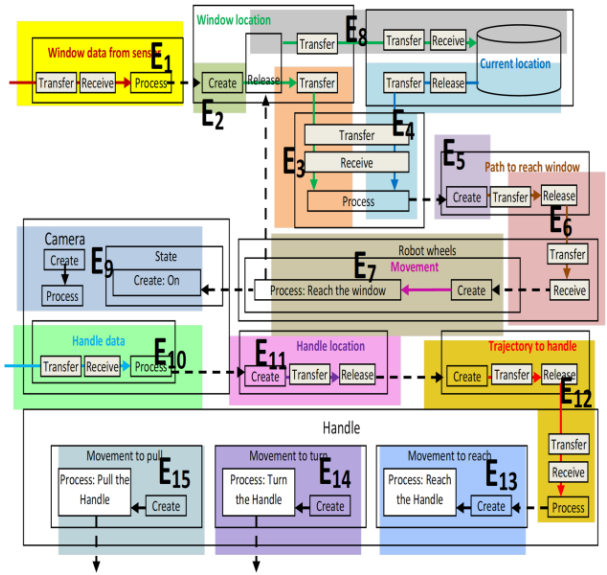

Fig. 11. Dividing the static model into parts.

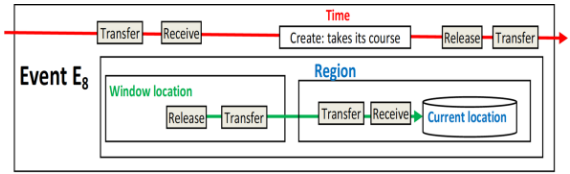

Fig. 12. Event E$_8$.

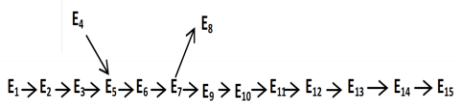

Fig. 13. The window-opening robot's behavior in terms of the chronology of events.

## VI. CASE STUDY

For our study, without loss of generality, we selected one architectural description for a robot called the NAO robot, the first autonomous, programmable humanoid robot created by SoftBank Group. It is an effective programming tool used in education and research. In addition, companies and health care centers might use it to welcome, inform, and entertain visitors [36]. NAO's documentation and user guide show how **to start the robot** and describe the result of turning the robot ON. In addition, they describe what happens when someone **approaches the robot** [37]. Furthermore, the robot's actions can be created and modified using the Choregraphe software. Choregraphe is a multiplatform desktop application that allows users to create animations and dialogues for robots. It also permits users to monitor and control the robot [36]. In addition, Figures 14 and 15 show the general architecture of **interacting with the robot** using Choregraphe software and Microsoft Azure, respectively [38]. Microsoft Azure is a continuously expanding cloud-computing service that allows users to build, manage, and deploy applications on a global network using their preferred tools and frameworks [39]. Our case study involves combining NAO's manual switching/approaching of the robot with the architecture in Figures 14 and 15. The result is demonstrated using a TM model to obtain an A-to-Z architecture of the robot.

The basic description of a user interacting with the robot is described below: [38]
1. Check the Internet connection.
2. Notify the users if NAO is in online mode.
3. Start speech recognition; by default, NAO will record a sound file when speech is detected.
4. The file path of the recorded sound clip will be sent to the Bing Speech API to extract text.
5. The text will then be sent to Azure Function to process the response.
6. In Azure Function, first, we will analyze the sentiment of the input, which will trigger a negative response (e.g., "Please don't scold me") when it detects that the users are upset.
7. If not, it will call QnA Maker API to obtain a response.
8. If an answer is found, then Azure Function will output the response. Otherwise, it will be recorded in Table Storage for admin to update.
9. Both the input and output are archived in Table Storage for validation.
10. Keywords in the responses from Azure Function will trigger specific movements.
11. Repeat step 3.
12. At any point in time, if the head is tapped, it will stop the conversation [38].

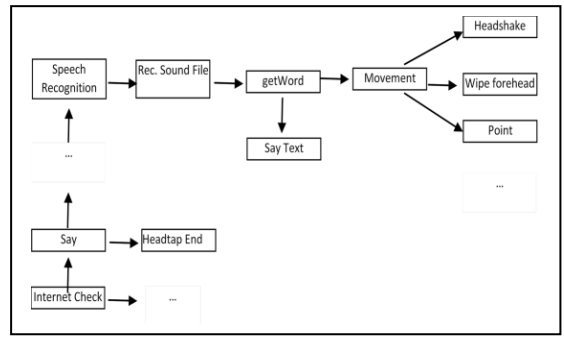

Fig. 14. An overall architecture of NAO robot using Choregraphe (partially from [38])

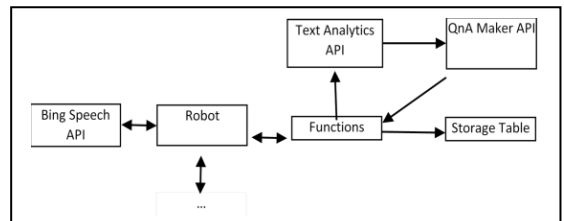

Fig. 15. An overall architecture using Azure (partially from [38])

*A. TM Static Model*

The robot consists of four parts: the sensors, the controller, the microphone, and the physical parts of the robot (head and body).

- Through the sensors, the user generates a signal (circle 1) that flows (2) to the robot to trigger the robot's state to be ON (3). Switching the robot ON triggers two actions: The LEDs are switched ON (4), and sound is created (5).
- When the LEDs are switched ON, the black-and-white light module creates (6) blinking, and a process (7) takes its course, changing color from black to white.
- Continuing at circle (5), a greeting sound is generated (8).
- Approaching the robot: A talking distance (assuming a maximum of 1.5 meters) is assumed to be initialized for the first time using the robot (9). When the user walks within this distance (10), a current distance is created (11) by the sensor, and its value flows (12) to the controller, where it is processed (13) to trigger (14) creation of the navy color (15) on the LEDs.
- Interacting with the robot: Once the user speaks (16—bottom-left corner of the figure), this act is detected (17) by the sensors and processed (18) to create (19) digital data from the analog sound. The digital data flow to the controller, where the data are processed to be recorded (20) and stored (21). Later, the stored data are retrieved and extracted as text (22) to create (23) sound clips. The sound clip is processed (24) to create (identify) its function (as a question or order).
- If the function is a question (25), it flows to be processed (26) such that:
    - The question is compared with the questions stored in the database.
        (i) If the question is found, the answer is retrieved (27) and flows to the microphone (28) to be processed (29). Accordingly, the corresponding speech is created (30).
        (ii) Otherwise, the question is stored in a database for validation (31).
- If the function is an order (32), the order flows to the physical head and body (33) to be processed (34). Based on the type of order, a physical action is performed as follows:
    I. Head shaking (35),
    II. Wiping forehead (36), or
    III. Pointing (37) in a certain direction.
- Interruption: At any moment, the user can tap the robot's head (38—bottom left), which generates a signal (39) from the tactile head sensors that flows (40) to the microphone. The signal is processed (41) to trigger stopping of the speech (42).

Figure 16 shows the TM model of the NAO robot's architecture.

*B. The Dynamic Model*

The decomposition of the S model forms the foundation upon which to understand events. The resulting parts of S should be sufficiently "meaningful." The meaningfulness of a part of S resides in the isomorphism between the part and the thing it is supposed to represent (in the modeler's conceptual framework). For example, "release" by itself as a sub-diagram does not seem to have this meaningfulness, but "release, transfer, transfer, and receive" is an ideal whole/part because it corresponds to the familiar notion of "moving from… to…" The resulting TM states (parts of S) are altered by inducing time (the time subthimac) to be transformed into events.

To construct the dynamic model, we identify the following events (see Figure 17):

Event 1 ($E_1$): The user presses the start button and creates a signal through the sensors.

Event 2 ($E_2$): The signal triggers the robot to be switched ON, which causes (i) the LEDs to blink, (ii) a greeting sound, and (iii) initialization of the talking distance.

Event 3 ($E_3$): The user approaches the robot within 1.5 m, which triggers creation of the current approaching distance.

Event 4 ($E_4$): The current distance flows to the controller.

Event 5 ($E_5$): The controller processes the current distance, and the LED light changes to navy.

Event 6 ($E_6$): The user speaks, which is received by the sensor.

Event 7 ($E_7$): The analog sound is processed and converted to digital data.

Event 8 ($E_8$): The digital data are released to the controller, where the data are stored.

Event 9 ($E_9$): The digital data are retrieved and extracted as text, then processed to create sound clips.

Event 10 ($E_{10}$): The sound clips are processed to trigger creation (identification) of the function.

Event 11 ($E_{11}$): The function is processed to distinguish a question from an order.

Event 12 ($E_{12}$): The function is a question, which is sent to the Q&A module.

Event 13 ($E_{13}$): The answer to the question is sent to the microphone.

Event 14 ($E_{14}$): The answer cannot be found; hence, it is stored.

Event 15 ($E_{15}$): The function is an order; hence, it is sent to the control of the physical body and head.

Event 16 ($E_{16}$): The order is processed.

Event 17 ($E_{17}$): The order is for the robot to shake its head.

Event 18 ($E_{18}$): The order is for the robot to wipe its forehead.

Event 19 ($E_{19}$): The order is for the robot to point.

Event 20 ($E_{20}$): The user taps the robot's head, creating a signal that is received by the microphone, thus stopping the sound.

Lastly, the robot's behavior can be specified by the chronology of events shown in Figure 18.

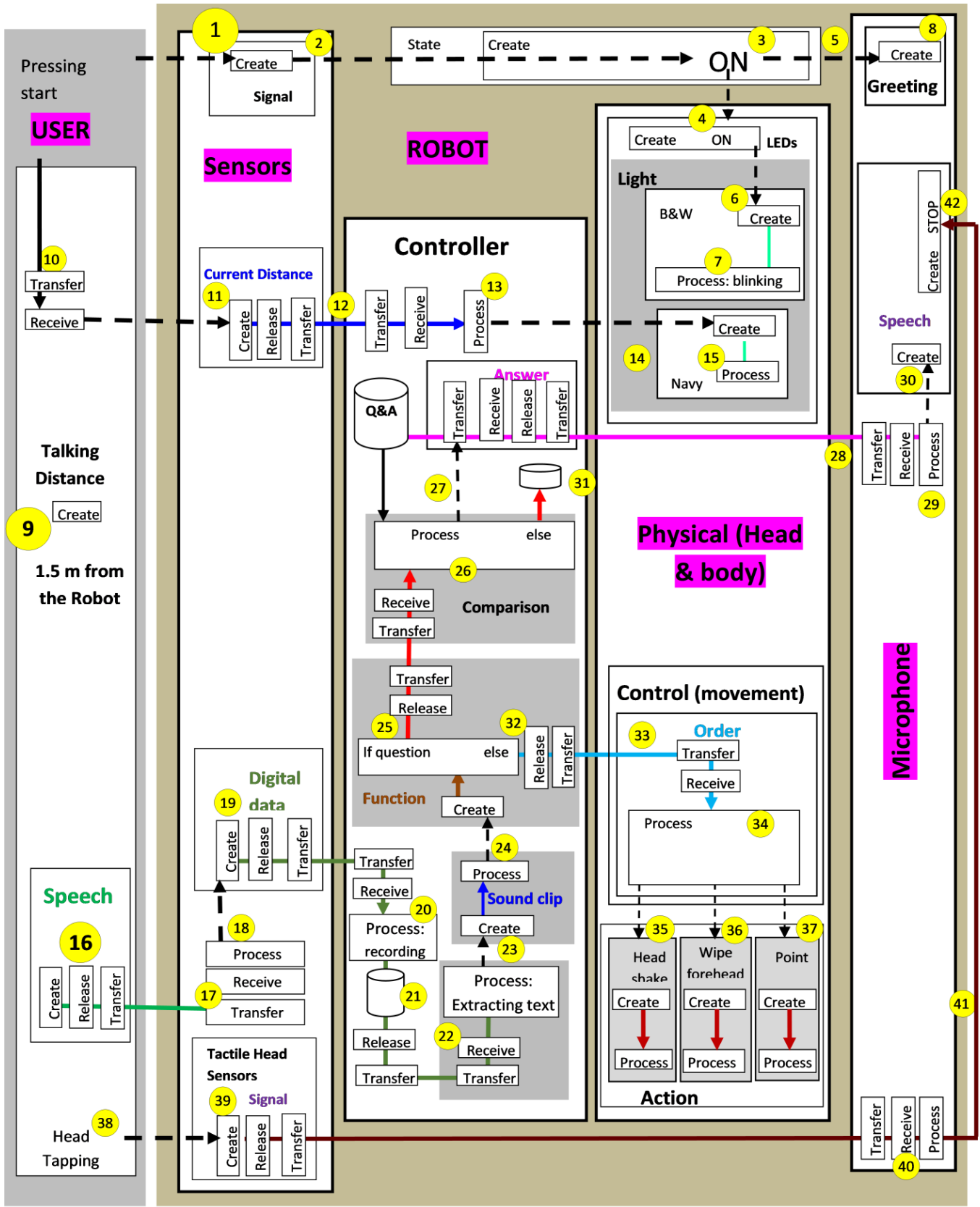

Fig. 16. TM model of the NAO robot's architecture.

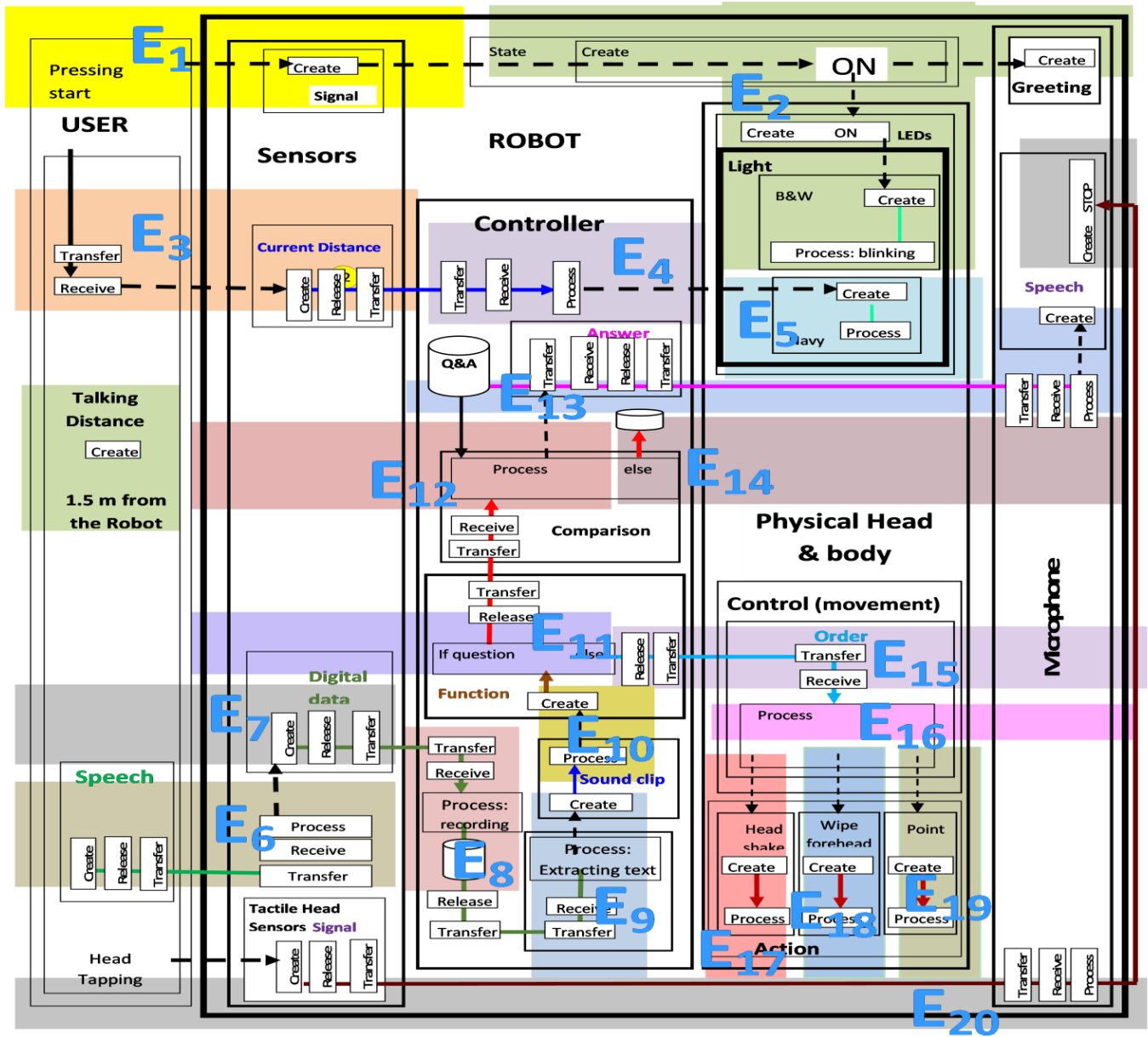

Fig. 17. The selected events

VII. CONCLUSION

This paper contributes to establishing a broad foundation for describing a high-level specification of robot systems. This involved developing the system, from static modeling to identifying the system's behavior. Our approach to present the benefits of such an approach was to contrast current architectural descriptions (Figures 1-4 and 14-15) with static and dynamic TM modeling to obtain a more detailed structure of the robot's processes, which is important to complete the designing and implementing phases of any robot structure. Further research will apply TM modeling to different aspects in robotics.

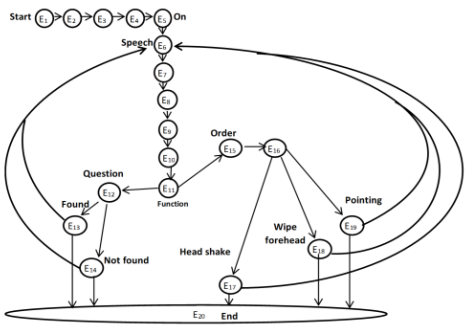

Fig. 18. The general architecture of interacting with the robot (from [38]).